\documentclass[10pt,twocolumn,letterpaper]{article}

\usepackage{cvpr}
\usepackage{times}
\usepackage{epsfig}
\usepackage{graphicx}
\usepackage{caption}
\usepackage{amsmath}
\usepackage{amssymb}
\usepackage{multirow}

\usepackage[pagebackref=true,breaklinks=true,letterpaper=true,colorlinks,bookmarks=false]{hyperref}

\usepackage{times}
\usepackage{epsfig}
\usepackage{graphicx}
\usepackage{float}
\usepackage{wrapfig}
\usepackage{amsmath,amssymb,amsthm}
\usepackage{algorithm,algorithmicx,algpseudocode}
\usepackage{bm,xspace}
\usepackage{comment}
\usepackage{verbatim}
\usepackage{multirow}
\usepackage{balance}
\usepackage{url}
\usepackage{booktabs}
\usepackage{etoolbox,siunitx}
\usepackage{calc}
\usepackage{pifont,hologo}
\usepackage[usenames, dvipsnames]{color}
\usepackage{nicefrac}

\newcommand{\ra}[1]{\renewcommand{\arraystretch}{#1}}

\usepackage[super]{nth}
\usepackage{nicefrac}
\robustify\bfseries

\DeclareMathOperator*{\argmin}{arg\,min}

\DeclareMathSymbol{@}{\mathord}{letters}{"3B}

\newcommand\mypara[1]{\vspace{1mm}\noindent\textbf{#1}}

\def\latex/{\LaTeX}
\def\bibtex/{\hologo{BibTeX}}

\cvprfinalcopy 

\def\cvprPaperID{1597} 

\ifcvprfinal\fi
\begin{document}

\title{Zoom to Learn, Learn to Zoom}

\author{Xuaner Zhang\\
UC Berkeley\\
\and
Qifeng Chen\\
HKUST\\
\and
Ren Ng\\
UC Berkeley\\
\and
Vladlen Koltun\\
Intel Labs\\
}

\twocolumn[{
\renewcommand\twocolumn[1][]{#1}%
\maketitle
\begin{center}
    \centering
    \begin{tabular}{@{}c@{\hspace{1mm}}c@{\hspace{1mm}}c@{\hspace{1mm}}c@{}}
    \includegraphics[width=.294\textwidth]{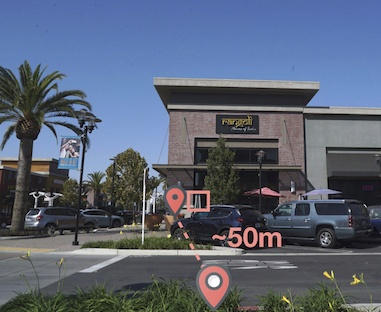} &
    \includegraphics[width=.23\textwidth]{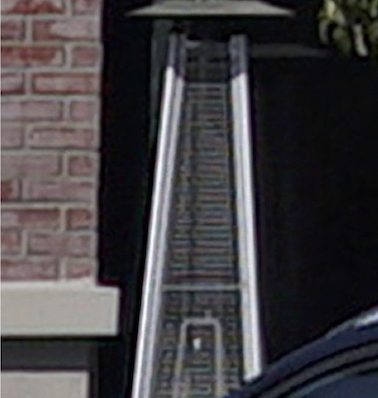} &
    \includegraphics[width=.23\textwidth]{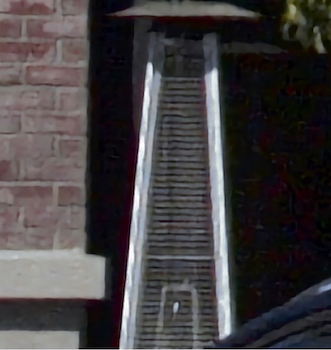} &
    \includegraphics[width=.23\textwidth]{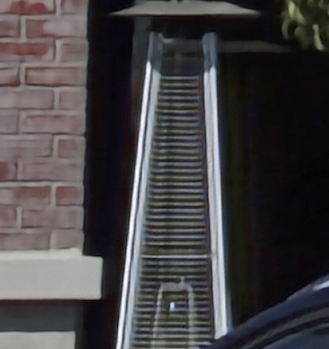}\\
    Input with distant object & ESRGAN & Ours-syn-raw & Ours\\
    \includegraphics[width=.294\textwidth]{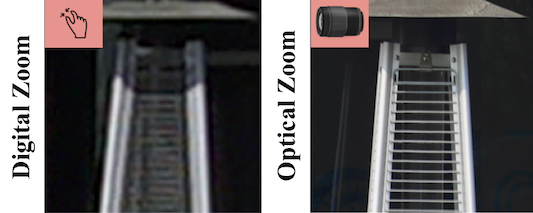} &
    \includegraphics[width=.23\textwidth]{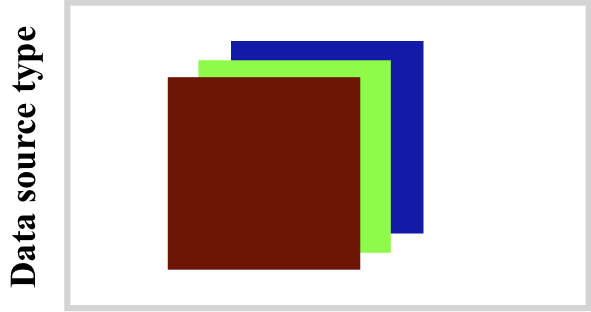} &
    \includegraphics[width=.23\textwidth]{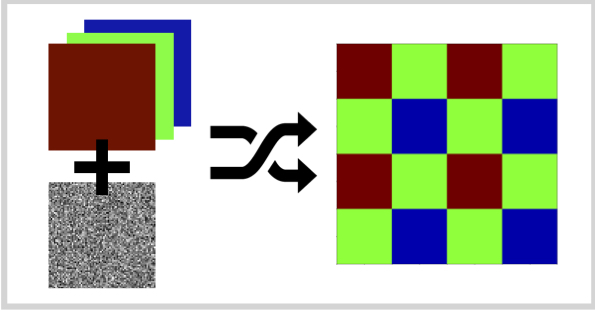} &
    \includegraphics[width=.23\textwidth]{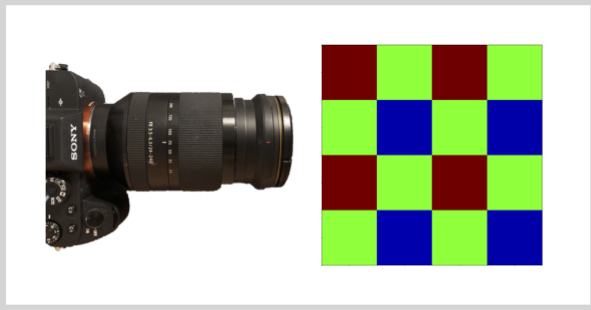}\\
    (A) Bicubic and ground truth & (B) 8-bit RGB & (C) Synthetic sensor & (D) Real sensor\\
    \end{tabular}
    \captionof{figure}{Our model (D) trained with real raw sensor data achieves better 4X computational zoom. We compare zoomed output against (B) ESRGAN~\cite{wang2018esrgan}, representative of state-of-the-art learning-based super-resolution methods, which operate on processed 8-bit RGB input, and (C) our model trained on synthetic sensor data. In (A), digital zoom via bicubic upsampling is the na\"{\i}ve baseline and optical zoom serves as the reference ground truth. Our output is artifact-free and preserves detail even for challenging regions such as the high-frequency grillwork.}
     \label{fig:teaser}
\end{center}}]


\begin{abstract}

This paper shows that when applying machine learning to digital zoom, it is beneficial to operate on real, RAW sensor data. Existing learning-based super-resolution methods do not use real sensor data, instead operating on processed RGB images. We show that these approaches forfeit detail and accuracy that can be gained by operating on raw data, particularly when zooming in on distant objects. The key barrier to using real sensor data for training is that ground-truth high-resolution imagery is missing. We show how to obtain such ground-truth data via optical zoom and contribute a dataset, SR-RAW, for real-world computational zoom. We use SR-RAW to train a deep network with a novel contextual bilateral loss that is robust to mild misalignment between input and outputs images. The trained network achieves state-of-the-art performance in 4X and 8X computational zoom. We also show that synthesizing sensor data by resampling high-resolution RGB images is an oversimplified approximation of real sensor data and noise, resulting in worse image quality.\footnote{Project website at: \url{https://ceciliavision.github.io/project-pages/project-zoom.html}}

\end{abstract}

\section{Introduction}
Zoom functionality is a necessity for mobile phones and cameras today. People zoom onto distant subjects such as wild animals and sports players in their captured images to view the subject in more detail. Smartphones such as iPhoneX are even equipped with two cameras at different zoom levels, indicating the importance of high-quality zoom functionality for the consumer camera market.

Optical zoom is an optimal choice for image zoom and can preserve high image quality, but zoom lenses are usually expensive and bulky. Alternatively, we can conveniently use digital zoom with a standard lens. However, digital zoom simply upsamples a cropped region of the camera sensor input, producing blurry output. It remains a challenge to obtain high-quality images for distant objects without expensive optical equipment.

We propose to improve the quality of super-resolution by starting with real raw sensor data. Recently, single-image super-resolution has progressed with deep models and learned image priors from large-scale datasets~\cite{bruna2015super,Johnson2016,kim2016deeply,Kim2016,Lai2017,ledig2017photo,lim2017enhanced,sajjadi2016enhancenet,zhang2018residual}. However, these methods are constrained in the following two respects. First, they approach computational zoom under a synthetic setup where the input image is a downsampled version of the high-resolution image, indirectly reducing the noise level in the input. In practice, regions of distant objects often contain more noise as fewer photons enter the aperture during the exposure time. Second, most existing methods start with an 8-bit RGB image that has been processed by the camera's image signal processor (ISP), which trades off high-frequency signal in higher-bit raw sensor data for other objectives (\eg noise reduction).

In this work, we raise the possibility to apply machine learning to computational zoom that uses real raw sensor data as input. The fundamental challenge is obtaining ground truth for this task: low-resolution raw sensor data with corresponding high-resolution images. One approach is to synthesize sensor data from 8-bit RGB images that are passed through some synthetic noise model~\cite{gharbi2016deep}. However, noise from a real sensor~\cite{tian2000noise} can be very challenging to model and is not modeled well by any current work that synthesizes sensor data for training. The reason is that sensor noise comes from a variety of sources, exhibiting color cross-talk and effects of micro-geometry and micro-optics close to the sensor surface. We find that while a model trained on synthetic sensor data works better than using 8-bit RGB data (\eg compare (B) and (C) in Figure~\ref{fig:teaser}), the model trained on real raw sensor data performs best (\eg compare (C) and (D) in Figure~\ref{fig:teaser}).

To enable learning from real raw sensor data for better computational zoom, we propose to capture real data with a zoom lens~\cite{kingslake1960development}, where the lens can move physically further from the image sensor to gather photons from a narrower solid angle for optical magnification. We build SR-RAW, the first dataset used for real-world computational zoom. \mbox{SR-RAW} contains ground-truth high-resolution images taken with high optical zoom levels. During training, an 8-bit image taken with a longer focal length serves as the ground truth for the higher-bit (\eg 12-14 bit) raw sensor image taken with a shorter focal length.

During training, SR-RAW brings up a new challenge: the source and target images are not perfectly aligned as they are taken with different camera configurations that cause mild perspective change. Furthermore, preprocessing introduces ambiguity in alignment between low- and high-resolution images. Mildly misaligned input-output image pairs make pixel-wise loss functions unsuitable for training. We thus introduce a novel contextual bilateral loss (CoBi) that is robust to such mild misalignment. CoBi draws inspiration from the recently proposed contextual loss (CX)~\cite{mechrez2018contextual}. A direct application of CX to our task yields strong artifacts because CX doesn't take spatial structure into account. To address this, CoBi prioritizes local features while also allowing for global search when features are not aligned.

In brief, we ``Zoom to Learn" -- collecting a dataset with ground-truth high-resolution images obtained via optical zoom, to ``Learn to Zoom" -- training a deep model that achieves better computational zoom. To evaluate our approach, we compare against existing super-resolution methods and also against an identical model to ours, but trained on synthetic sensor data obtained via a standard synthetic sensor approximation. Image quality is measured by distortion metrics such as SSIM, PSNR, and a learned perceptual metric. We also collect human judgments to validate the consistency of the generated images with human perception. Results show that real raw sensor data contains useful image signal for recovering high-fidelity super-resolved images. Our contributions can be summarized as follows:
\begin{itemize}
\item We demonstrate the utility of using real high-bit sensor data for computational zoom, rather than processed \mbox{8-bit} RGB images or synthetic sensor models.
\item We introduce a new dataset, SR-RAW, the first dataset for super-resolution from raw data, with optical ground truth. SR-RAW is taken with a zoom lens. Images taken with long focal length serve as optical ground truth for images taken with shorter focal length.
\item We propose a novel contextual bilateral loss (CoBi) that handles slightly misaligned image pairs. CoBi considers local contextual similarities with weighted spatial awareness.
\end{itemize}

\section{Related Work}
\mypara{Image Super-resolution.}
Image super-resolution has advanced from traditional filtering to learning-based methods. The goal is to reconstruct a high-resolution image from a low-resolution RGB image. Traditional approaches include filtering-based techniques such as bicubic upsampling and edge-preserving filtering~\cite{Li2001}. These filtering methods usually produce overly smooth texture in the output high-resolution image. Several approaches use patch matching to search for similar patches in a training dataset or in the image itself~\cite{Freeman2002,Glasner2009,Huang2015}. Recently, deep neural networks have been applied to super-resolution, trained with a variety of losses~\cite{Dong2016,Johnson2016,Kim2016}.

Many recent super-resolution approaches are based on generative adversarial networks. SRGAN~\cite{ledig2017photo} is an image super-resolution approach that applies a GAN to generate high-resolution images. The loss used in SRGAN combines a deep feature matching loss and an adversarial loss. Lai \etal.~\cite{Lai2017} propose the Laplacian Pyramid
Super-Resolution Network to progressively predict the residual of high-frequency details of a lower-resolution image in a coarse-to-fine image pyramid. Wang \etal.~\cite{wang2018esrgan} propose ESRGAN, which enhances image super-resolution with a Relativistic GAN~\cite{Jolicoeur_Martineau2019} that estimates how much one image is relatively more realistic than another. Wang \etal.~\cite{wang2018sftgan} study class-conditioned image super-resolution and propose SFT-GAN that is trained with a GAN loss and a perceptual loss. Most existing super-resolution models take a synthetic low-resolution RGB image (usually downsampled from a high-resolution image) as input. In contrast, we obtain real low-resolution images taken with shorter focal lengths and use optically zoomed images as ground truth.

\mypara{Image Processing with Raw Data.}
Prior works have used raw sensor data to enhance image processing tasks. Farsiua \etal.~\cite{farsiu2006multiframe} propose a maximum a posteriori technique for joint multi-frame demosaicing and super-resolution estimation with raw sensor data. Gharbi \etal.~\cite{gharbi2016deep} train a deep neural network for joint demosaicing and denoising. Zhou \etal.~\cite{zhou2018deep} address joint demosaicing, denoising, and super-resolution. These methods use synthetic Bayer mosaics. Similarly, Mildenhall \etal.~\cite{mildenhall2018burst} synthesize raw burst sequences for denoising. Chen \etal.~\cite{Chen2018} present a learning-based image processing pipeline for extreme low-light photography using raw sensor data. DeepISP is an end-to-end deep learning model that enhances the traditional camera image signal processing pipeline~\cite{Schwartz2019}. Similarly, we operate on raw sensor data and propose a method to super-resolve images by jointly optimizing for the camera image processing pipeline and super-resolution from raw sensor data.

\section{Dataset With Optical Zoom Sequences}
\label{sec:dataset}
\begin{figure*}
\centering
\begin{tabular}{ccc}
\multicolumn{3}{@{}c@{}}{\includegraphics[width=\textwidth]{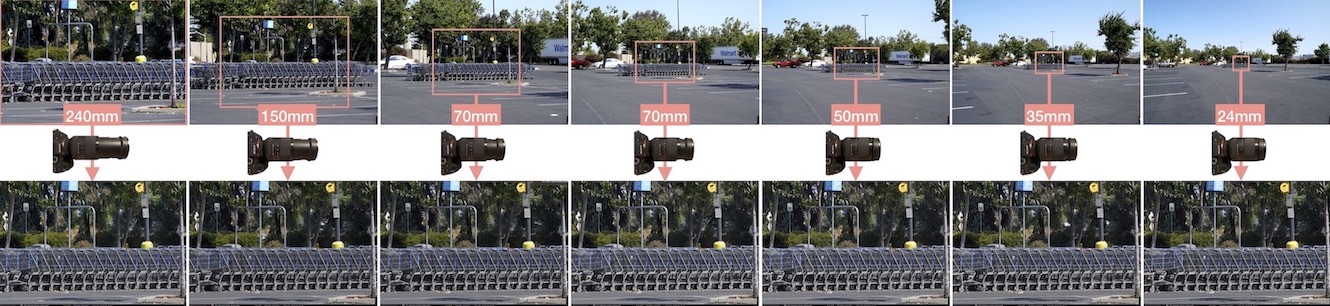}}\\
&(A) Example sequence from SR-RAW&\\
\multicolumn{3}{@{}c@{}}{\includegraphics[width=\textwidth]{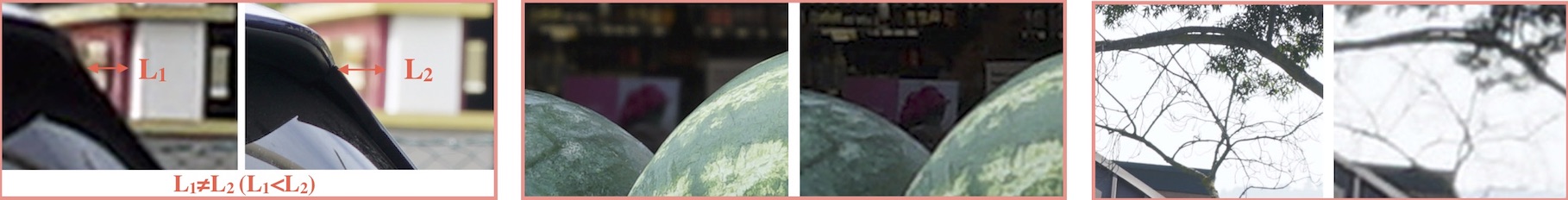}} \\
(B1) Noticeable perspective misalignment & (B2) Depth-of-field misalignment & (B3) Resolution alignment ambiguity\\
\end{tabular}
\caption{Example sequence from SR-RAW and three sources of misalignment in data capture and preprocessing. The unavoidable misalignment motivates our proposed loss.}
\label{fig:dataset}
\end{figure*}
To enable training with real raw sensor data for computational zoom, we collect a diverse dataset, SR-RAW, that contains raw sensor data and ground-truth high-resolution images taken with a zoom lens at various zoom levels. For data preprocessing, we align the captured images with different zoom levels via field of view (FOV) matching and geometric transformation. The SR-RAW dataset enables training an end-to-end model that jointly performs demosaicing, denoising, and super-resolution on raw sensor data. Training on real sensor data differentiates our framework from existing image super-resolution algorithms that operate on low-bit RGB images.

\subsection{Data Capture with a Zoom Lens}
\label{subsec:capture}
We use a 24-240 mm zoom lens to collect pairs of RAW images with different levels of optical zoom. Each pair of images forms an input-output pair for training a model: the short-focal-length raw sensor image is used as input and the long-focal-length RGB image is regarded as the ground-truth for super-resolution. For example, the RGB image taken with a 70mm focal length serves as the 2X zoom ground truth for the raw sensor data taken with a 35mm focal length. In practice, we collect 7 images under 7 optical zoom settings per scene for data collection efficiency. Every pair of images from the 7-image sequence forms a data pair for training a particular zoom model. In total, we collect 500 sequences in indoor and outdoor scenes. ISO ranges from 100 to 400. One example sequence is shown in Figure~\ref{fig:dataset}A.

During data capture, camera settings are important. First, depth of field (DOF) changes with focal length and it is not practical to adjust aperture size for each focal length to make DOF identical. We choose a small aperture size (at least f/20) to minimize the DOF difference (still noticeable in Figure~\ref{fig:dataset} B2), using a tripod to capture indoor scenes with a long exposure time. Second, we use the same exposure time for all images in a sequence so that noise level is not affected by focal length change. But we still observe noticeable illumination variations due to shutter and physical pupil being mechanical and involving action variation. This color variation is another motivation for us to avoid using pixel-to-pixel losses for training. Third, although perspective does not change with focal length, there exists slight variation (length of the lens) in the center of projection when the lens zooms in and out, generating noticeable perspective change between objects at different depths (Figure~\ref{fig:dataset} B1). Sony FE 24-240mm, the lens we use, requires a distance of at least 56.4 meters from the subject to have less than one-pixel perspective shift between objects that are 5 meters apart. Therefore, we avoid capturing very close objects but allow for such perspective shifts in our dataset.

\subsection{Data Preprocessing}
For a pair of training images, we denote the low-resolution image by RGB-L and its sensor data by RAW-L. For high-resolution ground truth we use RGB-H and RAW-H. We first match the field of view (FOV) between RAW-L and RGB-H. Alignment is then computed between RGB-L and RGB-H to account for slight camera movement caused by manually zooming the camera to adjust focal lengths. We apply a Euclidean motion model that allows image rotation and translation via enhanced correlation coefficient minimization~\cite{evangelidis2008parametric}. During training, RAW-L with matched FOV is fed into the network as input; its ground truth target is RGB-H that is aligned and has the same FOV with RAW-L. A scale offset is applied to the image if the optical zoom does not perfectly match the target zoom ratio. For example, an offset of 1.07 is applied to the target image if we use (35mm, 150mm) to train a 4X zoom model.

\subsection{Misalignment Analysis}
Misalignment is unavoidable during data capture and can hardly be eliminated by the preprocessing step. Since we capture data with different focal lengths, misalignment is inherently caused by the perspective changes as described in Section~\ref{subsec:capture}. Furthermore, when aligning images with different resolutions, sharp edges in the high-resolution image cannot be exactly aligned with blurry edges in the low-resolution image (Figure~\ref{fig:dataset} B3). The described misalignment in SR-RAW usually causes 40-80 pixel shifts in an 8-megapixel image pair.

\section{Contextual Bilateral Loss}
\label{sec:cobi}
\begin{figure*}
\begin{tabular}{@{\hspace{1mm}}c@{\hspace{1mm}}c@{\hspace{1mm}}c@{\hspace{1mm}}c@{}}
\includegraphics[width=.239\linewidth]{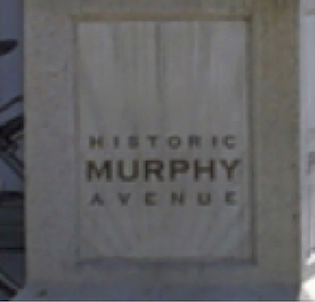} &
\includegraphics[width=.24\linewidth]{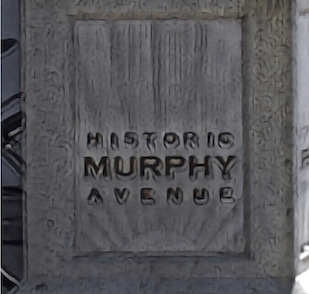} &
\includegraphics[width=.24\linewidth]{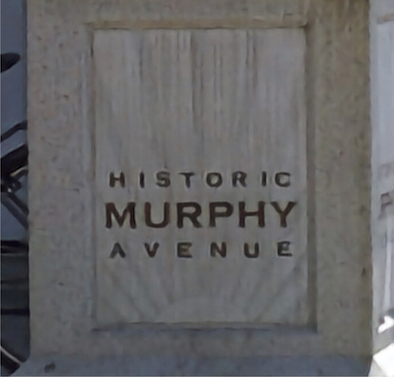} &
\includegraphics[width=.238\linewidth]{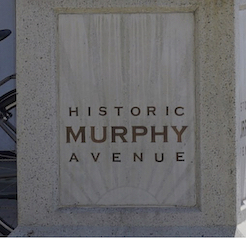}\\
(A) Bicubic & (B) Train with CX & (C) Train with CoBi & (D) Ground truth \\
\end{tabular}
\caption{Training with the contextual loss (CX) results in periodic artifacts as shown on the flat wall in (B). These artifacts are caused by inappropriate feature matching between source and target images, which does not take spatial location into account. In contrast, training with the proposed contextual bilateral loss (CoBi) leads to cleaner and better results, as shown in (C).}
\vspace*{-2mm}
\label{fig:artifact}
\end{figure*}

When using SR-RAW for training, we find that pixel-to-pixel losses such as $L_1$ and $L_2$ generate blurred images due to misalignment in the training data (Section~\ref{sec:dataset}). On the other hand, the recently proposed Contextual Loss (CX)~\cite{mechrez2018contextual} for unaligned data is also unsatisfactory as it only considers features but not their spatial location in the image. For a brief review, the contextual loss was proposed to train with unaligned data pairs. It treats the source image $P$ as a collection of feature points ${p_i}_{i=1}^{N}$ and the target image $Q$ as a set of feature points ${q_j}_{j=1}^{M}$. For each source image feature $p$, it searches for the nearest neighbor (NN) feature match $q$ such that $q=\argmin_{q}\mathbb{D}(p,q_j)_{j=1}^{M}$ under some distance measure $\mathbb{D}(p,q)$. Given input image $P$ and its target $Q$, the contextual loss tries to minimize the summed distance of all matched feature pairs, formulated as
\begin{equation}
\label{eq:cx}
\mathrm{CX}(P, Q)=\frac{1}{N}\sum_{i}^{N}\min_{j=1,...,M}(\mathbb{D}_{p_i,q_j}).
\end{equation}
We find that training with the contextual loss yields images that suffer from significant artifacts, demonstrated in Figure~\ref{fig:artifact}. We hypothesize that these artifacts are caused by inaccurate feature matching in the contextual loss. We thus analyze the percentage of features that are matched uniquely (i.e., bijectively). The percentage of target features matched with a unique source feature is only 43.7\%, much less than the ideal percentage of 100\%.

In order to train our model appropriately, we need to design an image similarity measure applicable to image pairs with mild misalignment. Inspired by the edge-preserving bilateral filter~\cite{tomasi1998bilateral}, we integrate the spatial pixel coordinates and pixel-level RGB information into the image features. Our Contextual Bilateral loss (CoBi) is defined as
\begin{equation}
\label{eq:cobi}
\mathrm{CoBi}(P, Q)=\frac{1}{N}\sum_{i}^{N}\min_{j=1,...,M}(\mathbb{D}_{p_i,q_j}+w_{s}\mathbb{D'}_{p_i,q_j}),
\end{equation}
where $\mathbb{D'}_{p_i,q_j}=\|(x_i,y_i) - (x_j,y_j)\|_2$. $(x_i,y_i)$ and $(x_j,y_j)$ are spatial coordinates of features $p_i$ and $q_j$, respectively, and $w_s$ denotes the weight of spatial awareness for nearest neighbor search. $w_s$ enables CoBi to be flexible to the amount of misalignment in the training dataset. The average number of one-to-one feature matches for our model trained with CoBi increases from 43.7\% to 93.9\%.

We experiment with different feature spaces for CoBi and conclude that a combination of RGB image patches and pre-trained perceptual features leads to the best performance. In particular, we use pretrained VGG-19 features \cite{Simonyan2015} and select `conv1\_2', `conv2\_2', and `conv3\_2' as our deep features, shown to be successful for image synthesis and enhancement \cite{chen2017photographic,zhang2018single}. Cosine distance is used to measure feature similarity. Our final loss function is defined as
\begin{equation}
\label{eq:obj}
\begin{aligned}
 \mathrm{CoBi_{RGB}}(P, Q, n)+\lambda\mathrm{CoBi_{VGG}}(P, Q),
\end{aligned}
\end{equation}
where we use $n \times n$ RGB patches as features for $\mathrm{CoBi_{RGB}}$, and $n$ should be larger for the 8X zoom (optimal $n=15$) than the 4X zoom model (optimal $n=10$). Qualitative comparisons on the effect of $\lambda$ are shown in the supplement.

\section{Experimental Setup}
\label{sec:experiments}
We use images from SR-RAW to train a 4X model and an 8X model. We pack each $2\times 2$ block in the raw Bayer mosaic into 4 channels as input for our model. The packing reduces the spatial resolution of the image by a factor of two in width and height, without any loss of signal. We subtract the black level and then normalize the data to $[0,1]$. White balance is read from EXIF metadata and applied to the network output as post-processing for comparison against ground truth. We adopt a 16-layer ResNet architecture~\cite{He2016} followed by $\log_2{N}+1$ up-convolution layers where $N$ is the zoom factor.

We split 500 sequences in SR-RAW into training, validation, and test sets with a ratio of 80:10:10, so that there are 400 sequences for training, 50 for validation, and 50 for testing. For a 4X zoom model, we get 3 input-output pairs per sequence for training, and for an 8X zoom model, we get 1 pair per sequence. Each pair contains a full-resolution (8-megapixel) Bayer mosaic image and its corresponding full-resolution optically zoomed RGB image. We randomly crop $64 \times 64$ patches from a full-resolution Bayer mosaic as input for training. Example training patches are shown in the supplement.

We first compare our approach to existing super-resolution methods that operate on processed RGB images. Then we conduct controlled experiments on our model variants trained on different source data types. All comparisons are tested on the 50 held-out test sequences from SR-RAW.

\subsection{Baselines}
\label{subsec:processed}
We choose a few representative super-resolution (SR) methods for comparisons: SRGAN~\cite{ledig2017photo}, a GAN-based SR model; SRResnet~\cite{ledig2017photo} and LapSRN~\cite{Lai2017}, which demonstrate different network architectures for SR; a model by Johnson \etal.~\cite{Johnson2016} that adopts perceptual losses; and finally ESRGAN~\cite{wang2018esrgan}, the winner of the most recent Perceptual SR Challenge PIRM~\cite{blau20182018}.

For all baselines except~\cite{Johnson2016}, we use public pretrained models; we first try to fine-tune their models on SR-RAW, adopting the standard setup in the literature: for each image, the input is the downsampled (bicubic) version of the target high-resolution image. However, we notice little difference in average performance ($<\!\!\pm$0.04 for SSIM, $<\!\!\pm$0.05 for PSNR, and $<\!\!\pm$0.025 for LPIPS) in comparison to the pretrained models without fine-tuning, and thus we directly use the models without fine-tuning for comparisons. For baseline methods without pretrained models, we train their models from scratch on SR-RAW.

\subsection{Controlled Experiments on Our Model}
\label{subsec:control}
\begingroup
\begin{figure}
\begin{center}
\renewcommand{\arraystretch}{1.42}
\begin{tabular}{@{}l@{\hspace{3mm}}l@{\hspace{1mm}}|l@{\hspace{1mm}}|l@{}}
\toprule
\multirow{5}{*}{\includegraphics[width=0.2\textwidth]{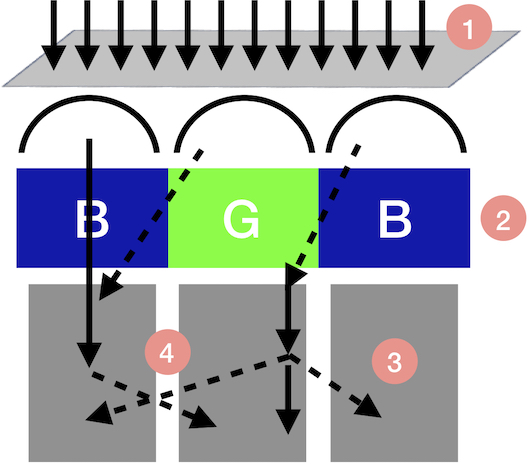}} & Features & Syn & Real \\
\cline{2-4}
& 1. AA Filter & No & Yes/No \\
& 2. Bit Depth & 8 & 12-14\\
& 3. Crosstalk & No & Yes\\
& 4. Fill Factor & 100\% & $<$100\%\\
\bottomrule
\end{tabular}
\setlength{\belowcaptionskip}{-20pt}
\caption{A range of sensor characteristics exist in real sensor data, but are not accurately reflected in synthesized sensor data. Each of the features listed in the table corresponds to its numbered label on the illustration, indicating the challenge to model realistic synthetic sensor data.}
\label{fig:sensor}
\end{center}
\end{figure}
\endgroup
\noindent\textbf{``Ours-png".}
For comparison, we also train a copy of our model (``Ours-png") using 8-bit processed RGB images to evaluate the benefits of having real raw sensor data. Different from the synthetic setup described in Section~\ref{subsec:processed}, instead of using downsampled RGB image as input, we use the RGB image taken with a shorter focal length as input. The RGB image taken with a longer focal length serves as the ground truth.
\newline\noindent\textbf{``Ours-syn-raw".}
To test whether synthesized raw data can replace real sensor data for training, we adopt the standard sensor synthesis model described by Gharbi et al.~\cite{gharbi2016deep} to generate synthetic Bayer mosaics from 8-bit RGB images. In brief, we retain one color channel per pixel according to the Bayer mosaic pattern from a white-balanced, gamma-corrected sRGB image, and introduce Gaussian noise with random variance. We train a copy of our model on these synthetic sensor data (``Ours-syn-raw") and test on real sensor data that is white-balanced and gamma-corrected.

\section{Results}
\subsection{Quantitative Evaluation}
To quantitatively evaluate the presented approach, we use the standard SSIM and PSNR metrics, as well as the recently proposed learned perceptual metric LPIPS~\cite{zhang2018unreasonable}, which measures perceptual image similarity using a pretrained deep network. Although there is mild misalignment in the input-output image pairs in SR-RAW (see Section~\ref{sec:dataset}), this misalignment exists across all methods and thus the comparisons are fair.

The results are reported in Table~\ref{table:quant}. They indicate that existing super-resolution models do not perform well on real low-resolution images that require digital zoom in practice. These models are trained under a synthetic setting where input images (usually downsampled) are clean and only contain 8-bit signal. GAN-based methods often generate noisy artifacts and lead to low PSNR and SSIM scores. Bicubic upsampling and SRResnet produce blurry results and get a low score in LPIPS. Our model, trained on high-bit real raw data and supervised by optically zoomed images, can effectively recover high-fidelity visual information with 4X and 8X computational zoom.

In Table~\ref{table:quant_control}, we show evaluations on our model trained with two different strategies. ``Ours-png" is our model trained on processed RGB images. By accessing real low-resolution data taken by a short focal length, the model learns to better handle noise, but its super-resolution power is limited by the low-bit image source. ``Ours-syn-raw" is our model trained on synthetic Bayer images. While the model gets access to raw sensor data during test time, it is limited by the domain gap between synthetic and real sensor data. We illustrate in Figure~\ref{fig:sensor} that a range of real sensor features are not reflected in a synthetic sensor model. Anti-aliasing filter (AA filter) exists in selected camera models. Synthetic sensor data is generated from 8-bit images while real sensor data contains high-bit signals. Inter-sensor crosstalk and sensor fill factor introduce noise into the color filter array and can be hardly parameterized by a simple noise model~\cite{yamashita2018intercolor}. The synthetic sensor model is insufficient to represent these complicated noise patterns.
\begin{table*}[ht]
\centering
\setlength{\tabcolsep}{3mm}
\ra{1.05}
\begin{tabular}{@{}l@{\hspace{10mm}}c@{\hspace{7mm}}c@{\hspace{7mm}}c@{\hspace{15mm}}c@{\hspace{7mm}}c@{\hspace{7mm}}c@{}}
\toprule
& & \textbf{4X} & & & \textbf{8X} & \\
\midrule
& SSIM$\uparrow$ & PSNR$\uparrow$ & LPIPS$\downarrow$ & SSIM$\uparrow$ & PSNR$\uparrow$ & LPIPS$\downarrow$\\
\midrule
Bicubic & 0.615 & 20.15 & 0.344 & 0.488 & 14.71 & 0.525  \\
SRGAN~\cite{ledig2017photo} & 0.384 & 20.31 & 0.260 & 0.393 & 19.23 & 0.395 \\
SRResnet~\cite{ledig2017photo} & 0.683 & 23.13 & 0.364 & 0.633 & 19.48 & 0.416\\
LapSRN~\cite{Lai2017} & 0.632 & 21.01 & 0.324 & 0.539 & 17.55 & 0.525 \\
Johnson \etal.~\cite{Johnson2016} & 0.354 & 18.83 & 0.270 & 0.421 & 18.18 & 0.394\\
ESRGAN~\cite{wang2018esrgan} & 0.603 & 22.12 & 0.311 & 0.662 & 20.68 & 0.416\\
\midrule
Ours & {\textbf{0.781}} & {\textbf{26.88}} & {\textbf{0.190}} & {\textbf{0.779}} & {\textbf{24.73}} & {\textbf{0.311}}\\
\bottomrule
\end{tabular}
\caption{Our model, trained with raw sensor data, performs better computational zoom than baseline methods, as measured by multiple metrics. Note that a lower LPIPS score indicates better image quality.}
\label{table:quant}
\end{table*}


\begin{table*}[ht]
\centering
\setlength{\tabcolsep}{3mm}
\ra{1.05}
\begin{tabular}{@{}l@{\hspace{10mm}}c@{\hspace{7mm}}c@{\hspace{7mm}}c@{\hspace{15mm}}c@{\hspace{7mm}}c@{\hspace{7mm}}c@{}}
\toprule
& & \textbf{4X} & & & \textbf{8X} & \\
\midrule
& SSIM$\uparrow$ & PSNR$\uparrow$ & LPIPS$\downarrow$ & SSIM$\uparrow$ & PSNR$\uparrow$ & LPIPS$\downarrow$\\
\midrule
Ours-png & 0.589 & 22.34 & 0.305 & 0.638 & 21.21 & 0.584 \\
Ours-syn-raw & {0.677} & {23.98} & {0.231} & {0.643} & {22.02} & {0.473} \\
\midrule
Ours & {\textbf{0.781}} & {\textbf{26.88}} & {\textbf{0.190}} & {\textbf{0.779}} & {\textbf{24.73}} & {\textbf{0.311}}\\
\bottomrule
\end{tabular}
\caption{Controlled experiments on our model, demonstrating the importance of using real sensor data.}
\label{table:quant_control}
\end{table*}

\subsection{Qualitative Results}
\begin{figure*}
\centering
\begin{tabular}{@{}c@{\hspace{1mm}}c@{\hspace{1mm}}c@{\hspace{1mm}}c@{\hspace{1mm}}c@{\hspace{1mm}}c@{}}
\includegraphics[width=3.5cm,height=3.6cm]{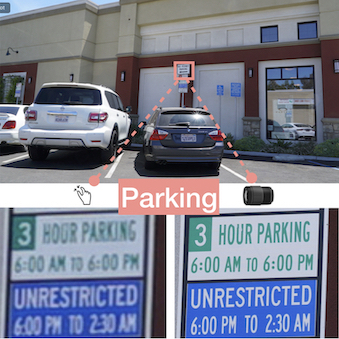} &
\includegraphics[width=2.3cm,height=3.6cm]{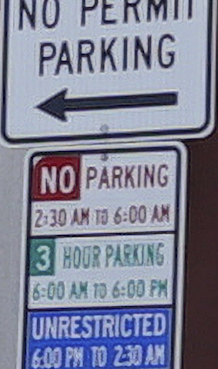} &
\includegraphics[width=2.3cm,height=3.6cm]{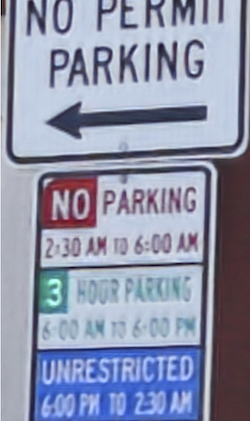} &
\includegraphics[width=2.3cm,height=3.6cm]{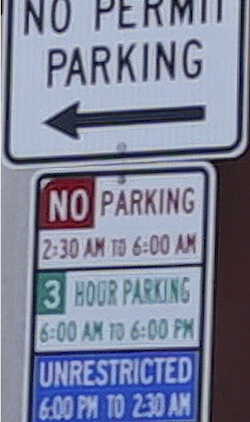} &
\includegraphics[width=2.3cm,height=3.6cm]{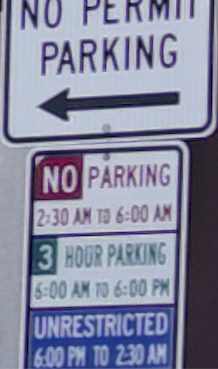} &
\includegraphics[width=2.3cm,height=3.6cm]{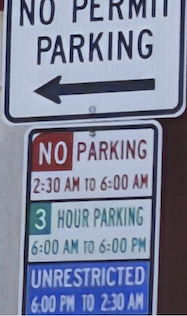} \\
\includegraphics[width=3.5cm,height=3.8cm]{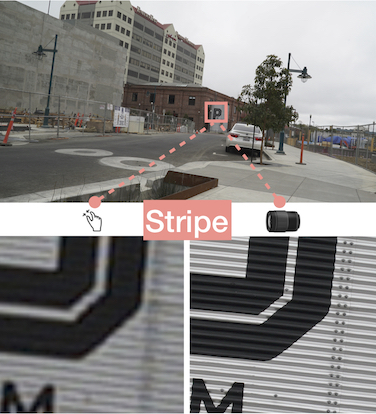} &
\includegraphics[width=2.3cm,height=3.8cm]{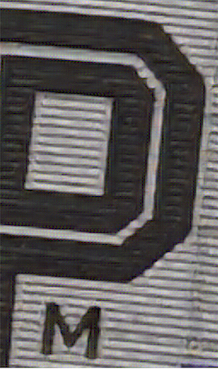} &
\includegraphics[width=2.3cm,height=3.8cm]{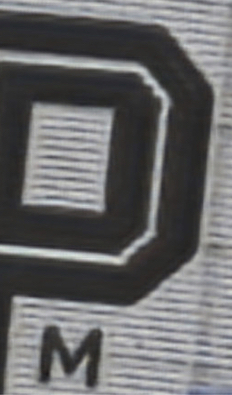} &
\includegraphics[width=2.3cm,height=3.8cm]{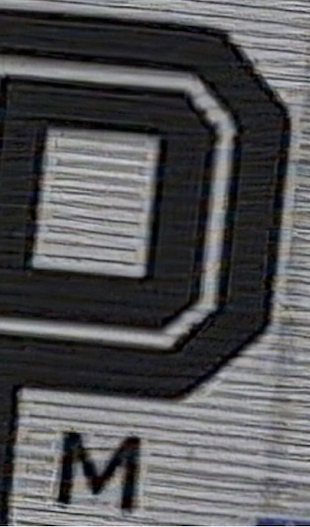} &
\includegraphics[width=2.3cm,height=3.8cm]{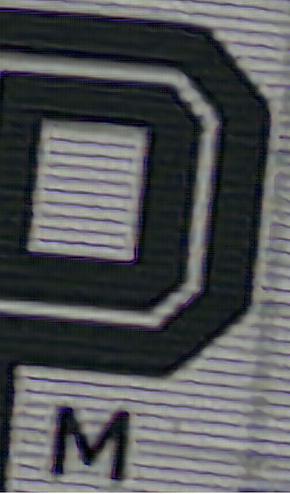} &
\includegraphics[width=2.3cm,height=3.8cm]{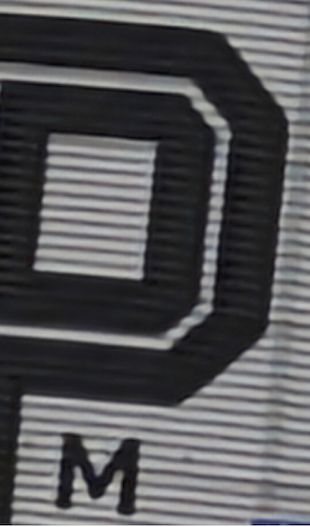} \\
\small \hspace{-1mm} Input \hspace{10mm} GT & \small Johnson \etal.~\cite{Johnson2016} &\small SRResnet~\cite{ledig2017photo} &  \small ESRGAN~\cite{wang2018esrgan} & \small LapSRN~\cite{Lai2017} & \small Ours\\
\end{tabular}
\caption{Our 4x zoom results show better perceptual performance in super-resolving distant objects against baseline methods that are trained under a synthetic setting and applied to processed RGB images.}
\label{fig:results_rgb}
\end{figure*}

\begin{figure*}[t]
\centering
\begin{tabular}{@{}c@{\hspace{1mm}}c@{\hspace{.8mm}}c@{\hspace{.8mm}}c@{\hspace{.8mm}}c@{}}
\includegraphics[width=4.5cm,height=3.3cm]{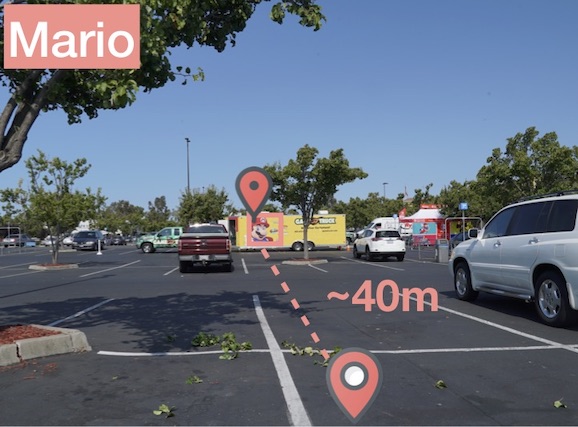} &
\includegraphics[width=2.8cm,height=3.3cm]{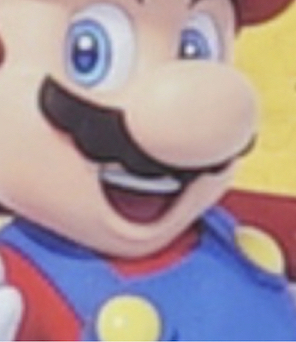} &
\includegraphics[width=2.8cm,height=3.3cm]{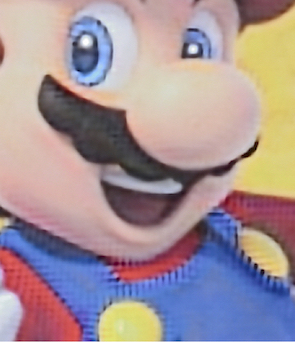} &
\includegraphics[width=2.8cm,height=3.3cm]{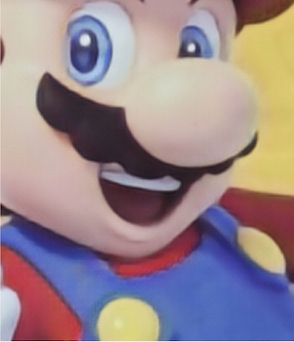} &
\includegraphics[width=2.8cm,height=3.3cm]{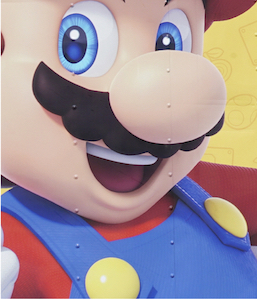} \\
\includegraphics[width=4.5cm,height=3.3cm]{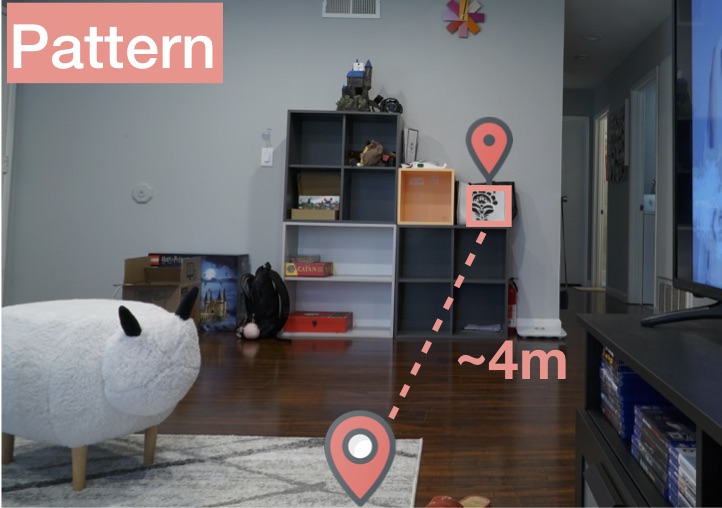} &
\includegraphics[width=2.8cm,height=3.3cm]{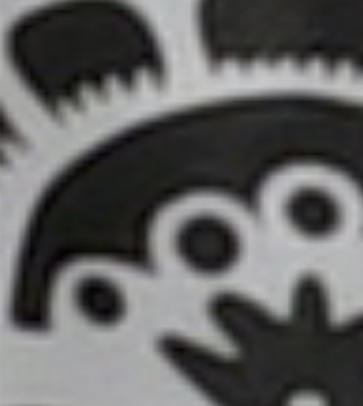} &
\includegraphics[width=2.8cm,height=3.3cm]{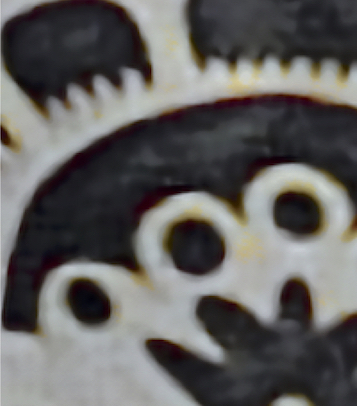} &
\includegraphics[width=2.8cm,height=3.3cm]{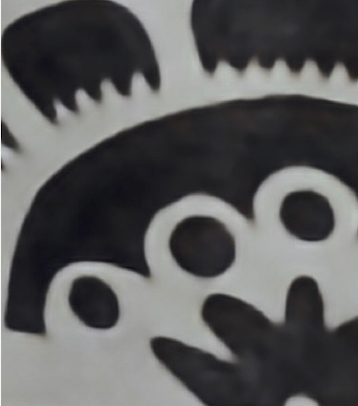} &
\includegraphics[width=2.8cm,height=3.3cm]{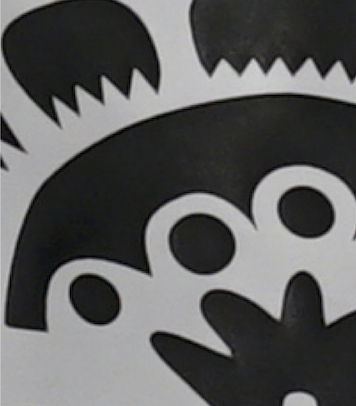} \\
\includegraphics[width=4.5cm,height=3.3cm]{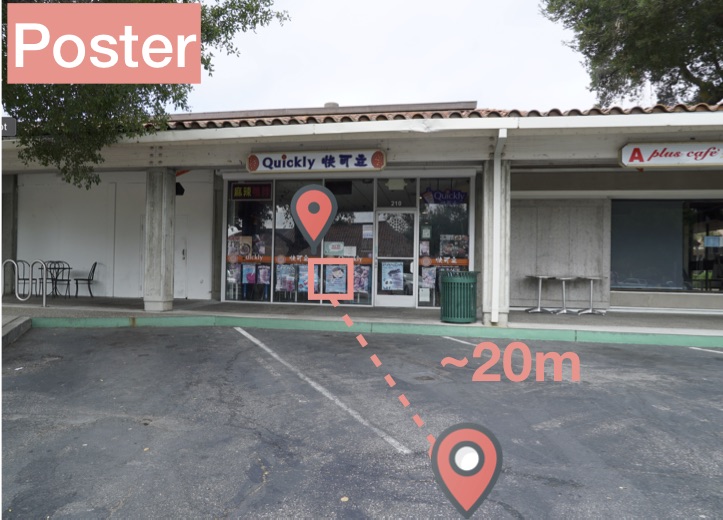} &
\includegraphics[width=2.8cm,height=3.3cm]{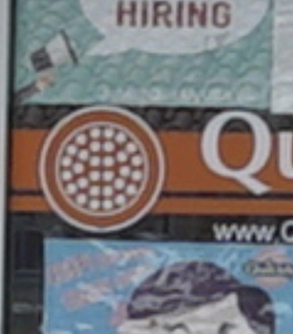} &
\includegraphics[width=2.8cm,height=3.3cm]{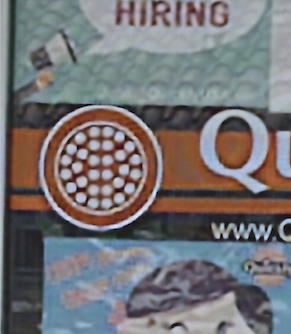} &
\includegraphics[width=2.8cm,height=3.3cm]{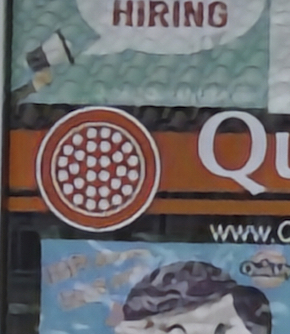} &
\includegraphics[width=2.8cm,height=3.3cm]{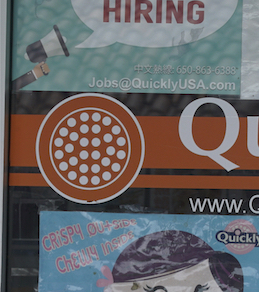} \\
Input & Bicubic & Synthetic sensor & Ours & GT\\
\end{tabular}
\caption{The model trained on synthetic sensor data produces artifacts such as jagged edges in ``Mario" and ``Poster" and color aberrations in ``Pattern", while our model, trained on real sensor data, produces clean and high-quality zoomed images.}
\label{fig:results_raw}
\end{figure*}

We show qualitative comparisons in Figure~\ref{fig:results_rgb} against baseline methods, and in Figure~\ref{fig:results_raw} against our model variants trained with different data. Most input images contain objects that are far from the viewpoint and require computational zoom in practice. Ground truth is obtained using a zoom lens with 4X optical zoom. In Figure~\ref{fig:results_rgb}, baseline methods fail to separate contents from the noise; it appears that their performance is limited by only having access to 8-bit signals in color images, especially in ``Stripe", which contains high-frequency details. Text in ``Parking" appear noisy in all baseline results, while our model generates a clean and discernible output image. In Figure~\ref{fig:results_raw}, the model trained on synthetic sensor data produces jagged edges in ``Mario" and ``Poster," and demosaic color artifacts in ``Pattern." Our model, trained on real sensor data with SR-RAW, can generate a clean demosaiced image with high image fidelity.

\subsection{Perceptual Experiments}
\label{subsec:percep}
We also evaluate the perceptual quality of our generated images by conducting a perceptual experiment on Amazon Mechanical Turk. In each task, we compare our model against a baseline on 100 4X-zoomed images (50 test images from SR-RAW and additional 50 images taken without ground truth). We conduct blind randomized A/B testing against LapSRN, Johnson \etal., ESRGAN, and our model trained on synthetic sensor data. We show the participants both results side by side, in random left-right order. The original low-resolution image is also presented for reference. We ask the question: ``A and B are two versions of the high-resolution image of the given low-resolution image. Which image (A or B) has better image quality?" In total, 50 workers participated in the experiment. The results, listed in Table~\ref{tab:user_study}, indicate that our model produces images that are seen as more realistic in a significant majority of blind pairwise comparisons.
\begin{table}
\centering
\setlength{\tabcolsep}{3mm}
\ra{1.15}
\begin{tabular}{@{}lc@{}}
\toprule
& Preference rate \\
\midrule
Ours$>$Syn-raw& 80.6\%\\
Ours$>$ESRGAN~\cite{wang2018esrgan} & 83.4\%\\
Ours$>$LapSRN~\cite{Lai2017} & 88.5\%\\
Ours$>$Johnson \etal.~\cite{Johnson2016} & 92.1\% \\
\bottomrule
\end{tabular}
\setlength{\belowcaptionskip}{-10pt}
\caption{Perceptual experiments show that our results are strongly preferred over baseline methods.}
\label{tab:user_study}
\end{table}

\subsection{Generalization to Other Sensors}
Different image sensors have different structural noise patterns in their Bayer mosaics (See Figure~\ref{fig:sensor}). Our model, trained on one type of Bayer mosaic, may not perform as well when applied to a Bayer mosaic from another device (\eg iPhoneX). To explore the potential of generalization to other sensors, we capture 50 additional iPhoneX-DSLR data pairs in outdoor environments. We fine-tune our model with only 5000 iterations to adapt our model to the iPhoneX sensor. A qualitative result is shown in Figure~\ref{fig:iphone} and more results can be found in the supplement. The results indicate that our pretrained model can be generalized to another sensor by fine-tuning the model on a small dataset captured with that sensor, and also indicate that input-output data pairs can come from different devices, suggesting the application of our method to devices with limited optical zoom power.
\begingroup
\begin{figure}[H]
\centering
\begin{tabular}{@{}c@{\hspace{0.8mm}}c@{\hspace{2mm}}c@{\hspace{0.8mm}}c@{}}
\multirow{2}{*}[9mm]{\rotatebox{90}{\hspace{2mm} Smartphone Input}} &
\multirow{2}{*}[10mm]{\includegraphics[width=0.43\linewidth]{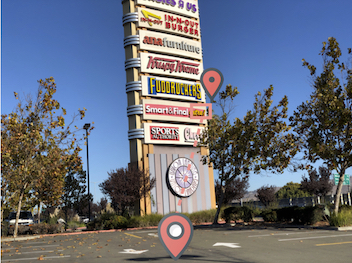}} &
\rotatebox{90}{\hspace{1mm} Bicubic} &
\includegraphics[width=0.43\linewidth]{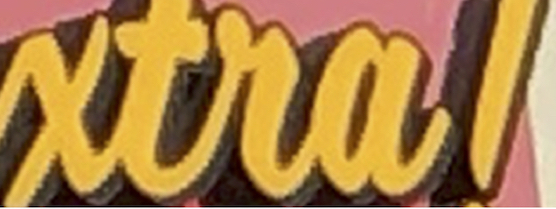} \\
& & \rotatebox{90}{\hspace{2mm} Ours} &
\includegraphics[width=0.43\linewidth]{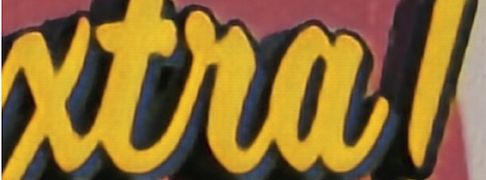} \\
\end{tabular}
\vspace*{-2mm}
\setlength{\belowcaptionskip}{-6pt}
\caption{Our model can adapt to input data from a different sensor after fine-tuning on a small dataset.}
\label{fig:iphone}
\end{figure}
\endgroup

\section{Conclusion}
We have demonstrated the effectiveness of using real raw sensor data for computational zoom. Images are directly super-resolved from raw sensor data via a learned deep model that performs joint ISP and super-resolution. Our approach absorbs useful signal from the raw data and produces higher-fidelity results than models trained on processed RGB images or synthetic sensor data. To enable training with real sensor data, we collect a new dataset that contains optically-zoomed images as ground truth and introduce a novel contextual bilateral loss that is robust to mild misalignment in training data pairs. Our results suggest that learned models could be integrated into cameras for high-quality digital zoom. Our work also indicates that preserving signal from raw sensor data may be beneficial for other image processing tasks.


{\small
\bibliographystyle{ieee}
\bibliography{main}

\begin{thebibliography}{10}\itemsep=-1pt

\bibitem{blau20182018}
Y.~Blau, R.~Mechrez, R.~Timofte, T.~Michaeli, and L.~Zelnik-Manor.
\newblock The 2018 {PIRM} challenge on perceptual image super-resolution.
\newblock In {\em ECCV Workshops}, 2018.

\bibitem{bruna2015super}
J.~Bruna, P.~Sprechmann, and Y.~LeCun.
\newblock Super-resolution with deep convolutional sufficient statistics.
\newblock In {\em ICLR}, 2015.

\bibitem{Chen2018}
C.~Chen, Q.~Chen, J.~Xu, and V.~Koltun.
\newblock Learning to see in the dark.
\newblock In {\em CVPR}, 2018.

\bibitem{chen2017photographic}
Q.~Chen and V.~Koltun.
\newblock Photographic image synthesis with cascaded refinement networks.
\newblock In {\em ICCV}, 2017.

\bibitem{Dong2016}
C.~Dong, C.~C. Loy, K.~He, and X.~Tang.
\newblock Image super-resolution using deep convolutional networks.
\newblock {\em {IEEE} Trans. Pattern Anal. Mach. Intell.}, 2016.

\bibitem{evangelidis2008parametric}
G.~D. Evangelidis and E.~Z. Psarakis.
\newblock Parametric image alignment using enhanced correlation coefficient
  maximization.
\newblock {\em {IEEE} Trans. Pattern Anal. Mach. Intell.}, 2008.

\bibitem{farsiu2006multiframe}
S.~Farsiu, M.~Elad, and P.~Milanfar.
\newblock Multiframe demosaicing and super-resolution of color images.
\newblock {\em {IEEE} Trans. Image Processing}, 2006.

\bibitem{Freeman2002}
W.~T. Freeman, T.~R. Jones, and E.~C. Pasztor.
\newblock Example-based super-resolution.
\newblock {\em {IEEE} Computer Graphics and Applications}, 2002.

\bibitem{gharbi2016deep}
M.~Gharbi, G.~Chaurasia, S.~Paris, and F.~Durand.
\newblock Deep joint demosaicking and denoising.
\newblock {\em ACM Trans. on Graphics (TOG)}, 2016.

\bibitem{Glasner2009}
D.~Glasner, S.~Bagon, and M.~Irani.
\newblock Super-resolution from a single image.
\newblock In {\em ICCV}, 2009.

\bibitem{He2016}
K.~He, X.~Zhang, S.~Ren, and J.~Sun.
\newblock Deep residual learning for image recognition.
\newblock In {\em CVPR}, 2016.

\bibitem{Huang2015}
J.~Huang, A.~Singh, and N.~Ahuja.
\newblock Single image super-resolution from transformed self-exemplars.
\newblock In {\em CVPR}, 2015.

\bibitem{Johnson2016}
J.~Johnson, A.~Alahi, and L.~Fei{-}Fei.
\newblock Perceptual losses for real-time style transfer and super-resolution.
\newblock In {\em ECCV}, 2016.

\bibitem{Jolicoeur_Martineau2019}
A.~Jolicoeur{-}Martineau.
\newblock The relativistic discriminator: A key element missing from standard
  {GAN}.
\newblock In {\em ICLR}, 2019.

\bibitem{kim2016deeply}
J.~Kim, J.~Kwon~Lee, and K.~Mu~Lee.
\newblock Deeply-recursive convolutional network for image super-resolution.
\newblock In {\em CVPR}, 2016.

\bibitem{Kim2016}
J.~Kim, J.~K. Lee, and K.~M. Lee.
\newblock Accurate image super-resolution using very deep convolutional
  networks.
\newblock In {\em CVPR}, 2016.

\bibitem{kingslake1960development}
R.~Kingslake.
\newblock The development of the zoom lens.
\newblock {\em Journal of the SMPTE}, 1960.

\bibitem{Lai2017}
W.-S. Lai, J.-B. Huang, N.~Ahuja, and M.-H. Yang.
\newblock Deep {Laplacian} pyramid networks for fast and accurate
  super-resolution.
\newblock In {\em CVPR}, 2017.

\bibitem{ledig2017photo}
C.~Ledig, L.~Theis, F.~Husz{\'a}r, J.~Caballero, A.~Cunningham, A.~Acosta,
  A.~P. Aitken, A.~Tejani, J.~Totz, Z.~Wang, and W.~Shi.
\newblock Photo-realistic single image super-resolution using a generative
  adversarial network.
\newblock In {\em CVPR}, 2017.

\bibitem{Li2001}
X.~Li and M.~T. Orchard.
\newblock New edge-directed interpolation.
\newblock {\em {IEEE} Trans. Image Processing}, 2001.

\bibitem{lim2017enhanced}
B.~Lim, S.~Son, H.~Kim, S.~Nah, and K.~M. Lee.
\newblock Enhanced deep residual networks for single image super-resolution.
\newblock In {\em CVPR Workshops}, 2017.

\bibitem{mechrez2018contextual}
R.~Mechrez, I.~Talmi, and L.~Zelnik-Manor.
\newblock The contextual loss for image transformation with non-aligned data.
\newblock In {\em ECCV}, 2018.

\bibitem{mildenhall2018burst}
B.~Mildenhall, J.~T. Barron, J.~Chen, D.~Sharlet, R.~Ng, and R.~Carroll.
\newblock Burst denoising with kernel prediction networks.
\newblock In {\em CVPR}, 2018.

\bibitem{sajjadi2016enhancenet}
M.~S. Sajjadi, B.~Sch{\"o}lkopf, and M.~Hirsch.
\newblock {EnhanceNet}: Single image super-resolution through automated texture
  synthesis.
\newblock In {\em ICCV}, 2017.

\bibitem{Schwartz2019}
E.~Schwartz, R.~Giryes, and A.~M. Bronstein.
\newblock {DeepISP}: Toward learning an end-to-end image processing pipeline.
\newblock {\em {IEEE} Trans. Image Processing}, 2019.

\bibitem{Simonyan2015}
K.~Simonyan and A.~Zisserman.
\newblock Very deep convolutional networks for large-scale image recognition.
\newblock In {\em ICLR}, 2015.

\bibitem{tian2000noise}
H.~Tian.
\newblock {\em Noise analysis in {CMOS} image sensors}.
\newblock PhD thesis, Stanford University, 2000.

\bibitem{tomasi1998bilateral}
C.~Tomasi and R.~Manduchi.
\newblock Bilateral filtering for gray and color images.
\newblock In {\em ICCV}, 1998.

\bibitem{wang2018sftgan}
X.~Wang, K.~Yu, C.~Dong, and C.~C. Loy.
\newblock Recovering realistic texture in image super-resolution by deep
  spatial feature transform.
\newblock In {\em CVPR}, 2018.

\bibitem{wang2018esrgan}
X.~Wang, K.~Yu, S.~Wu, J.~Gu, Y.~Liu, C.~Dong, C.~C. Loy, Y.~Qiao, and X.~Tang.
\newblock {ESRGAN}: Enhanced super-resolution generative adversarial networks.
\newblock In {\em ECCV}, 2018.

\bibitem{yamashita2018intercolor}
Y.~Yamashita and S.~Sugawa.
\newblock Intercolor-filter crosstalk model for image sensors with color filter
  array.
\newblock {\em IEEE Trans. on Electron Devices}, 2018.

\bibitem{zhang2018unreasonable}
R.~Zhang, A.~A. Efros, E.~Shechtman, and O.~Wang.
\newblock The unreasonable effectiveness of deep features as a perceptual
  metric.
\newblock In {\em CVPR}, 2018.

\bibitem{zhang2018single}
X.~Zhang, R.~Ng, and Q.~Chen.
\newblock Single image reflection separation with perceptual losses.
\newblock In {\em CVPR}, 2018.

\bibitem{zhang2018residual}
Y.~Zhang, Y.~Tian, Y.~Kong, B.~Zhong, and Y.~Fu.
\newblock Residual dense network for image super-resolution.
\newblock In {\em CVPR}, 2018.

\bibitem{zhou2018deep}
R.~Zhou, R.~Achanta, and S.~S{\"u}sstrunk.
\newblock Deep residual network for joint demosaicing and super-resolution.
\newblock In {\em Color and Imaging Conference}, 2018.

\end{thebibliography}
}

\end{document}